\documentclass[final]{cvpr}
\usepackage{xcolor}

\usepackage{times}
\usepackage{epsfig}
\usepackage{graphicx}
\usepackage{amsmath}
\usepackage{amssymb}
\usepackage{comment}

\usepackage[pagebackref=true,breaklinks=true,colorlinks,bookmarks=false]{hyperref}

\def\cvprPaperID{5443} 
\def\confYear{CVPR 2021}
\definecolor{amber}{rgb}{1.0, 0.75, 0.0}

\newif\ifcomments
\commentstrue 

\ifcomments
\newcommand{\LT}[1]{\textcolor{blue}{{\bf LT:} #1}}
\newcommand{\GB}[1]{\textcolor{red}{{\bf GB:} #1}}
\newcommand{\XL}[1]{\textcolor{purple}{{\bf XL:} #1}}
\newcommand{\JW}[1]{\textcolor{amber}{{\bf JW:} #1}}
\newcommand{\DP}[1]{\textcolor{blue}{{\bf DP:} #1}}
\else 
\newcommand{\LT}[1]{}
\newcommand{\GB}[1]{}
\newcommand{\XL}[1]{}
\newcommand{\JW}[1]{}
\newcommand{\DP}[1]{}
\fi

\begin{document}

\title{\textsc{Vx2Text}: End-to-End Learning of Video-Based Text Generation \\ From Multimodal Inputs}

\author{Xudong Lin$^{1}$
\and
Gedas Bertasius$^2$
\and
Jue Wang$^2$
\and 
Shih-Fu Chang$^{1}$
\and
Devi Parikh$^{2,3}$
\and 
Lorenzo Torresani$^{2,4}$
\\
$^{1}$Columbia University \hspace{30pt} $^{2}$Facebook AI \hspace{30pt} $^{3}$Georgia Tech \hspace{30pt} $^{4}$Dartmouth
\\
}

\maketitle

\begin{abstract}
We present \textsc{Vx2Text}, a framework for text generation from multimodal inputs consisting of video plus text, speech, or audio. In order to leverage transformer networks, which have been shown to be effective at modeling language, each modality is first converted into a set of language embeddings by a learnable tokenizer. This allows our approach to perform multimodal fusion in the language space, thus eliminating the need for ad-hoc cross-modal fusion modules. To address the non-differentiability of tokenization on continuous inputs (e.g., video or audio), we utilize a relaxation scheme that enables end-to-end training. Furthermore, unlike prior encoder-only models, our network includes an autoregressive decoder to generate open-ended text from the multimodal embeddings fused by the language encoder. This renders our approach fully generative and makes it directly applicable to different ``video+$x$ to text'' problems without the
need to design specialized network heads for each task. The proposed framework is not only conceptually simple but also remarkably effective: experiments demonstrate that our approach based on a single architecture outperforms the state-of-the-art on three video-based text-generation tasks---captioning, question answering and audio-visual scene-aware dialog. 

\end{abstract}

%
%
%
%
%
%
%
%

\section{Introduction}

Among the fundamental goals of AI is the development of conversational multimodal systems that can reliably perceive the real-world and communicate with humans in natural language. Progress in this area has been dramatically advanced in recent years by the introduction of large-scale benchmarks assessing the ability to interpret audiovisual information and translate this understanding to natural language. Prime examples include datasets for image or video captioning~\cite{MSCOCO, Flickr30K, MSRVTTXuEtal:CVPR2016, KrishnaEtAl:ICCV2017, YouCook2ZhouEtAl:AAAI2018, lei2020tvr}, question answering (QA)~\cite{VQA:ICCV2015, GaoEtAl:NIPS2015, VMadlibs, Visual7W, Tgif-qa, MovieFIB, MovieQA, lei2018tvqa}, as well as audio-visual dialog~\cite{VisDial, AVSD}. In order to perform well on such benchmarks, the model must accomplish several goals: (1) extract salient information from each individual modality, (2) effectively combine the different cues to address the given query, and (3) generate and present the results in human-comprehensible text. 

In this paper, we present \textsc{Vx2Text}, a simple video-based approach that embeds these three steps in a unified, end-to-end trainable framework.  Objectives (1) and (2) are accomplished by utilizing modality-specific classifiers to convert the semantics from each input signal into a common semantic language space, which enables the application of powerful language models to directly interpret multimodal content. Specifically, our approach takes the textual labels of the top classes predicted by each classifier pretrained on existing datasets~\cite{Kinetics,gemmeke2017audio} and transforms them into word embeddings, using a pretrained language model~\cite{BERT,T5}. The benefit of this solution is that it opens up the possibility to carry out multimodal fusion by means of powerful language encoders such as T5~\cite{T5} without the need to design specialized cross-modal network modules~\cite{lu2019vilbert, li2020hero, ActBERT,luo2020univilm} or to resort to pretext tasks to learn to combine the different input signals~\cite{VideoBERT, ZhouEtAl:AAAI2020, li2020hero}. Not only is such a design much simpler but it also leads to better performance compared to prior approaches. 




In order to fulfill objective (3), we employ a generative text decoder~\cite{T5}, which transforms the multimodal features computed by the encoder into text, thus realizing the goal of generating results in human-comprehensible language. While prior multimodal works based on encoder-only architectures~\cite{VideoBERT,LXMERT,lu2019vilbert} are limited to operate in settings involving selection from a fixed set of text candidates, our generative approach can be used for open-ended sentence generation 
as, e.g., required in dialog applications. In addition, the use of a text decoder allows us to tackle different ``video+$x$ to text'' problems (e.g., answering and generating questions, dialog, as well as captioning) with the same architecture, without having to design specialized network heads for each task. 

We integrate these conceptually-distinct steps into a single architecture, which we train end-to-end. To achieve this, we adopt a differential tokenization on continuous modalities (e.g., audio or video) which renders the entire model---including the modality-specific classifiers---trainable with respect to the final objective. Our experiments demonstrate that our unified framework trained end-to-end produces significant performance gains over separately learned modules. Our  \textsc{Vx2Text} based on a single architecture trained in a generative fashion without any multimodal pretext pretraining outperforms the state-of-the-art on three different text-generation tasks---captioning, QA and dialog. 

\section{Related Work} 


Significant progress has been made in the area of vision and language, especially in the design of multimodal conversational agents that interact with humans in natural language, e.g., for question answering (QA)~\cite{VQA:ICCV2015, GaoEtAl:NIPS2015, VMadlibs, Visual7W, Tgif-qa, MovieFIB, MovieQA, lei2018tvqa} and audio-visual dialog~\cite{VisDial, AVSD}. Several approaches have been introduced for these tasks~\cite{lu2016hierarchical,anderson2018bottom,kottur2018visual,shah2019cycle,hori2019end,schwartz2019simple,le2019multimodal,yang2020bert,kim2020modality,lei2019tvqa+,li2020tmt,ZhouEtAl:AAAI2020}.

For example, Shah \etal~\cite{shah2019cycle} have proposed to leverage the cycle-consistency of question answering and question generation to improve the robustness of image question answering models on rephrased questions. Differently from our approach, the actual question and answer sentences are not decoded. Yang \etal~\cite{yang2020bert} have explored an encoder-only model using multimodal fusion of BERT representations and visual features for video QA~\cite{lei2018tvqa}. Being a discriminative approach, it is limited to selecting from the provided answer choices. Le \etal~\cite{le2019multimodal} proposed a multimodal attentional generative model, which fuses information from texts and audiovisual features and generates responses for audiovisual scene-aware dialog. While this and a few other recent works~\cite{ZhouEtAl:AAAI2020} have leveraged decoders for text-generation from multimodal inputs, we believe we are the first to empirically demonstrate via systematic ablations the performance improvements achieved with generative learning with decoding, compared to discriminative learning applied to the same encoder model. Furthermore, we note that the networks proposed in~\cite{le2019multimodal, ZhouEtAl:AAAI2020} include specialized cross-modal blocks, which as noted above, approach the task quite differently from our method. Experimental comparisons to these prior works show the superior performance of our design. 

There is also a family of multimodal transformer-based models~\cite{VideoBERT,ZhouEtAl:AAAI2020,li2020hero,LXMERT,lu2019vilbert} inspired by the success of pretext tasks in the language domain~\cite{BERT,radford2018improving,T5}. These works rely on expensive pretext training on large-scale datasets to learn multimodal representations. However, our \textsc{Vx2Text} can perform multimodal fusion in a unified language space, which does not require multimodal pretext training. 

We note that we are not the first to propose using labels of categories recognized from the audiovisual channels as input to language models. For example, detected object labels have been employed for image captioning~\cite{YinOrdonez:2017, AndersonEtAl:CVPR2018} and also video QA~\cite{lei2018tvqa}. However, differently from these prior works, we adopt a differentiable tokenization on continuous modalities which makes the entire model trainable end-to-end with respect to the final objective. Our experiments demonstrate the performance benefits of our approach.


\begin{figure*}[t]
\begin{center}
\includegraphics[width=0.95\linewidth]{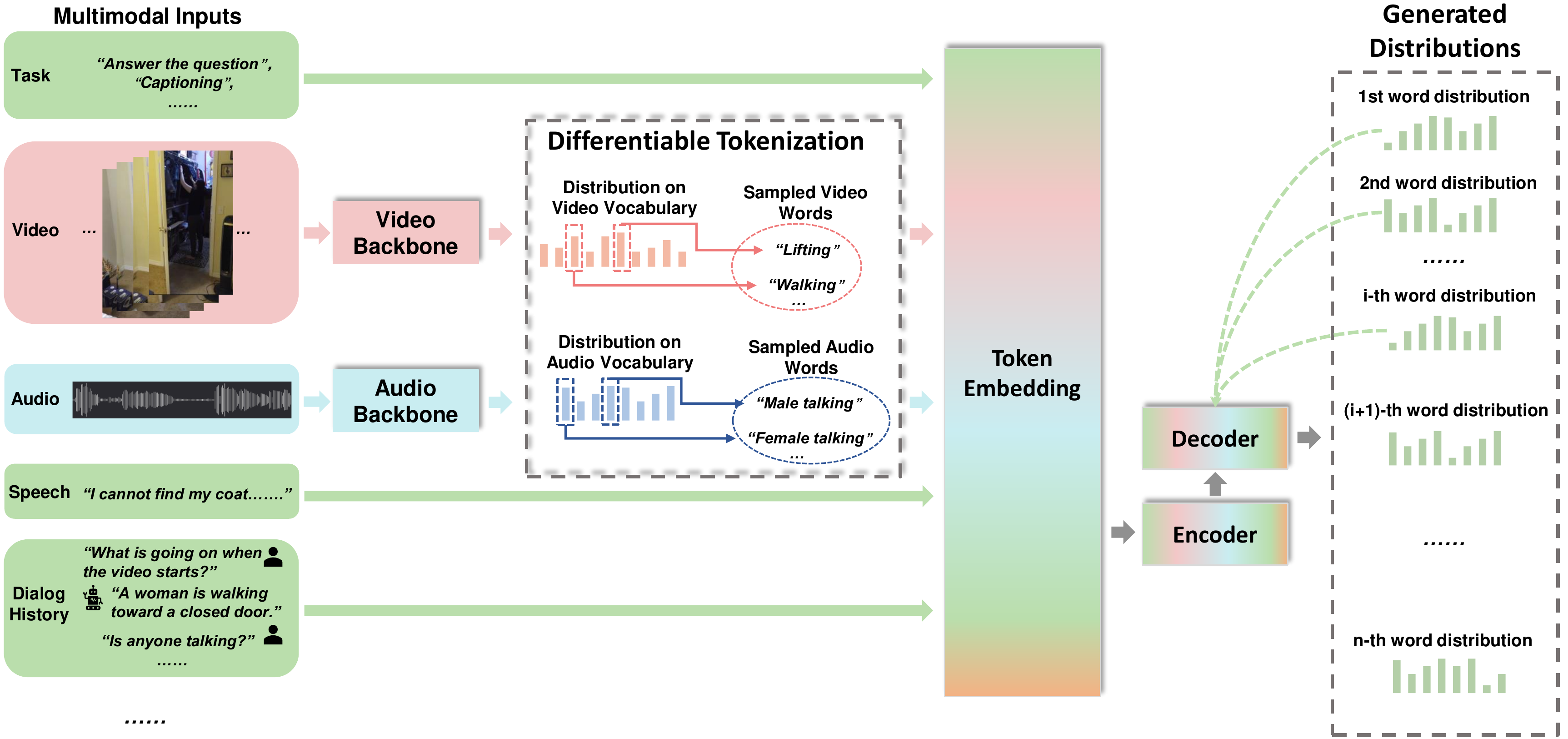}
\end{center}
\caption{Illustration of our proposed framework. \textsc{Vx2Text} receives as input a task specifier, and video with accompanying modalities, such as audio and speech. Each modality is converted into a set of tokens by means of modality-specific classifiers and a differentiable tokenization scheme that enables end-to-end training. Finally, an encoder-decoder architecture performs multimodal fusion in the language space and generates as output open-ended text addressing the given task.}
\label{illustration}
\end{figure*}

\section{Technical Approach}

Our goal is to design a unified framework that can generate open-ended text from video and accompanying modalities, e.g., audio, speech, or dialog history. We are specifically interested in tasks such as video captioning, question answering and audio-visual scene-aware dialog.  

Formally, let $\mathbf{x} = \{\mathbf{x}_1, \mathbf{x}_2, ..., \mathbf{x}_{M}\}$ be a multimodal sample, where $\mathbf{x}_m$ denotes the $m$-th modality. We specify the task that we want our model to address using a special task token $t\in \{Answer, Caption, dialog,...\}$. Our goal is then to train a model $\mathcal{F}(t, \mathbf{x}_1, \mathbf{x}_2, ..., \mathbf{x}_{M};\mathbf{W})$ that generates a sequence of text tokens $\mathbf{y}=[\mathbf{y}_1, \mathbf{y}_2, ..., \mathbf{y}_{N}]$ representing the output for task $t$. $\mathbf{W}$ denotes its trainable parameters. Depending on the task, our generated text may be in the form of answers, questions, interactive responses in a dialog, or captions. 

At a high-level, our approach can be summarized in three steps. First, we leverage pretrained modality-specific classifiers to obtain most probable category predictions for each modality. We then embed the textual names of the predicted categories into a semantic language space via our proposed differentiable tokenization scheme, which enables end-to-end training of the whole system including the modality-specific classifiers. Finally, we employ a generative encoder-decoder language model~\cite{T5} for mapping the embedding vectors from the multiple modalities into free-form text. This allows us to reformulate different ``video+$x$ to text'' problems as a single sequence-to-sequence task. We now present each of these steps in more detail. 

\subsection{Differentiable Tokenization}

Most prior methods~\cite{le2019multimodal,yang2020bert,li2020hero} rely on extra cross-modal fusion modules for combining input signals from different modalities. This renders the integration of different modalities burdensome and computational costly. Instead, we propose to perform multimodal fusion by mapping the different input signals into a common semantic language space through a simple scheme. We first leverage modality-specific classifiers trained to predict a large set of categories over predefined language vocabularies. These include video models trained to recognize a large collection of actions~\cite{Kinetics}, or audio classifiers distinguishing a broad set of sound categories~\cite{gemmeke2017audio}. Afterwards, we can utilize existing language embedding models to map the top textual categories predicted by each modality-specific classifier into a common semantic language space. 

Although conceptually simple, this approach has a few weaknesses. First, the pretrained modality-specific classifiers may not generalize to the target data. Second, the selection of the top categories from each classifier is not differentiable and thus prevents us from finetuning the modality-specific classifiers with respect to our target task. To address these limitations, we propose a differentiable tokenization scheme, which enables end-to-end training of the whole system including the modality-specific classifiers.

Let us denote with $\{\mathcal{N}_1, \mathcal{N}_2, ..., \mathcal{N}_{M}\}$ a set of modality-specific networks. For each modality $m$ we use a network model  $\mathcal{N}_m$ pretrained for a classification task on a predefined category space $\mathcal{C}_m = \{1, ..., C_m\}$. Let $p_{m}(c|\mathbf{x}) \in [0, 1]$ be the normalized probabilistic output of $\mathcal{N}_m(\mathbf{x}_m)$ for category $c \in \{1, ..., C_m\}$, such that $\sum_{c=1}^{C_m} p_{m}(c|\mathbf{x}) =1$. 
We convert these classification predictions into a set of text-embedding vectors by (1) sampling $K_m$ categories (without replacement) from the probabilistic outputs for each modality $m$ and then (2) embedding the names of the sampled categories via a matrix multiplication:

\begin{equation}
    \mathbf{e}^k_m = \mathbf{W}_{m}^T \mathbf{c}^k_m.
\end{equation}

where $\mathbf{W}_{m} \in \mathbb{R}^{C_m \times D}$ is a learned $D$-dimensional embedding of $C_m$ category tokens and $\mathbf{c}^k_m$ is a one-hot vector that encodes the name of the $k$-th sampled category from modality $m$.

Note that the sampling process is necessary during training because directly selecting top predictions will drop the rich information in the predicted distributions and bias the training process~\cite{jang2016categorical}. In order to make the sampling differentiable, we leverage the Gumbel-Softmax trick~\cite{jang2016categorical} and a differentiable approximation of tokenization~\cite{bengio2013estimating}. Specifically, we reparameterize the predicted probability distribution $ \mathbf{p}_m \in \mathcal{R}^{1 \times C_m}$ by adding Gumbel noise $ \mathbf{g}_m  \in \mathcal{R}^{1 \times C_m}$ to it, where $\mathbf{g}_m = -\log {(-\log{(\mathbf{u})})}$ with $\mathbf{u}\sim \text{Uniform}(0,1)$.  We then sample the top $K_m$ categories from the reparameterized distribution $\tilde{ \mathbf{p}}_m \in \mathcal{R}^{1 \times C_m}$ for each modality $m$.



With this re-parameterized distribution, selecting the top $K_m$ categories is equivalent to sampling $K_m$ categories from the original distribution. For detailed proof, we refer the reader to~\cite{kool2019stochastic}. However, the process of selecting the top $K_m$ categories is still not differentiable. To address this issue, we use a Straight-Through Estimator~\cite{jang2016categorical}. Specifically, during forward propagation, we sample top $K_m$ categories as described above. Instead, during backward propagation we estimate the gradient for each category $c$ as:

\begin{equation}
    \nabla_{} G \approx \nabla_{\mathbf{W}_m} \frac{\exp{(\log{p_m(c|\mathbf{x})} + \mathbf{g}_m(c))}}{\sum_{c' }^{|\mathcal{C}_m|} \exp{(\log{p_m(c'|\mathbf{x})} + \mathbf{g}_m(c'))}}.
\end{equation}




This leads to a unified formulation, which enables end-to-end learning of the entire system including the modality-specific classifiers. Furthermore, note that the embedding transformation $\mathbf{W}_m$ can be initialized using a pretrained language embedding space~\cite{T5}. This simple procedure provides the advantage of converting all modalities into the same semantic language space, thus eliminating the need for designing complex cross-modal fusion blocks. Furthermore, we can seamlessly leverage powerful language encoders for our target task, which is highly beneficial.



\subsection{Generative Encoder-Decoder}
With the different modalities embedded in the same language space, we can directly use a text encoder to fuse the multimodal information. We collect the embedding vector $\mathbf{e}_{t}$ representing the task definition $t$ together with the embeddings computed from the different modalities into a sequence of $L$ vectors which we feed into the text encoder $\mathcal{F}_{En}$:
\begin{equation}
    \mathbf{z} = \mathcal{F}_{En} (\mathbf{e}_{t}, \mathbf{e}_{\text{S}}, \mathbf{e}^1_1,..., \mathbf{e}^{K_1}_1,\mathbf{e}_{\text{S}},...,\mathbf{e}_{\text{S}},
    \\ \mathbf{e}^1_{M},..., \mathbf{e}^{K_M}_M),
\end{equation}
where $\mathbf{e}_{\text{S}}$ is the embedding of a special ``separator'' token, and $\mathbf{z}\in \mathbb{R}^{L\times d'}$ is a sequence of $L$ vectors with a dimensionality $d'$. 
The features $\mathbf{z}$ produced by the text encoder capture task-specific information from multiple modalities. 

Afterwards, we feed the new representation $\mathbf{z}$ to the decoder for text generation. Our decoder generates results in an auto-regressive manner, meaning that it uses previously decoded outputs as part of its input. Formally, we can write this as follows:
\begin{equation}
    \hat{\mathbf{g}}_i = \mathcal{F}_{De} (\mathbf{z}, \tilde{\mathbf{g}}_{1},..., \tilde{\mathbf{g}}_{i-1}),
\end{equation}
where $\hat{\mathbf{g}}_i  \in \mathbb{R}^{T'}$ is the $i$-th decoded distribution over a dictionary of $T'$ tokens and $\{\tilde{\mathbf{g}}_1,...,\tilde{\mathbf{g}}_{i-1}\}$ are history tokens. The decoding process will terminate when the ``End-of-Sequence'' token is generated.

\subsection{Training}

During training, we follow the common practice of teacher-forcing~\cite{williams1989learning, T5}, which means that we replace the decoding history with ground-truth tokens $\mathbf{g}_{i}$ in the corresponding positions:


\begin{equation}
    \hat{\mathbf{g}}_i = \mathcal{F}_{De} (\mathbf{z}, \mathbf{g}_{1},..., \mathbf{g}_{i-1}),
\end{equation}

Our entire system is then trained with a standard cross-entropy loss:

\begin{equation}
    \mathcal{L} = \min_{\mathbf{w}} \frac{1}{n} \sum_i \text{Cross-Entropy}(\hat{\mathbf{g}}_i, \mathbf{g}_i),
\end{equation}

where $n$ is the number of valid tokens. Note, that this design supports generation of text with variable length. While here we show the objective for a single training sample, in practice we optimize over mini-batches of samples. 


\subsection{Inference}

Most previous multimodal transformers~\cite{ZhouEtAl:AAAI2020,li2020hero} rely on task-specific heads to tackle different tasks. Specifically, the heads designed for generative tasks typically differ substantially from those used in discriminative settings.  However, our \textsc{Vx2Text} seamlessly addresses both types of tasks without the need to change its architecture.


For generative tasks, e.g., captioning and video dialog, we follow previous works and use Beam Search \cite{schwartz2019simple,le2019multimodal} (with beam width set to 5) or Greedy Decoding~\cite{lei2020tvr} to generate coherent sentences. Instead, for discriminative tasks, e.g., question answering on TVQA, the model is required to pick the most probable answer from a provided candidate set. In such cases, we include the entire set of candidate answers as additional input to the model (using separator tokens to mark them) and then evaluate each candidate output under the probability distribution defined by the autoregressive decoder. Finally, we select the highest-probability answer among the choices as the prediction.
In this way, with a unified encoder-decoder structure, our model can handle both generative and discriminative tasks. In our experiments we demonstrate that the knowledge stored in the decoder helps our generative \textsc{Vx2Text} outperform its discriminative counterpart as well as previous discriminative models (see Sections \ref{sec:gen} and \ref{sec:sota}).

\subsection{Implementation Details}

We use R(2+1)D-34 \cite{tran2018closer,ghadiyaram2019large} trained on Kinetics \cite{Kinetics} as our video backbone network, with the 400 action categories of Kinetics as the video vocabulary. We follow the video preprocessing procedure described in~\cite{ghadiyaram2019large}: during training, we randomly sample a clip of 32 frames; during testing, we uniformly sample 10 clips and construct a pool of predictions. We sample $K_v = 12$ predicted categories from the pool to represent the action/events in videos. Note that the sampled sequence is ordered temporally for predictions from different clips; predictions from the same clip are ordered according to the confidence score.

As the audio backbone, we use CNN14 \cite{kong2020panns}, which is trained on AudioSet \cite{gemmeke2017audio} to recognize 527 acoustic events. Audio segments are sampled at 16,000 Hz from corresponding video clips and then are processed to extract Log-mel spectrograms which are fed into the CNN. We use $K_a = 6$ predicted categories to represent the acoustic events in audio segments.
We provide analyses on the hyperparameters $K_v$ and $K_a$ in the appendix.

\begin{figure*}[!t]
\begin{center}
\includegraphics[width=0.95\linewidth]{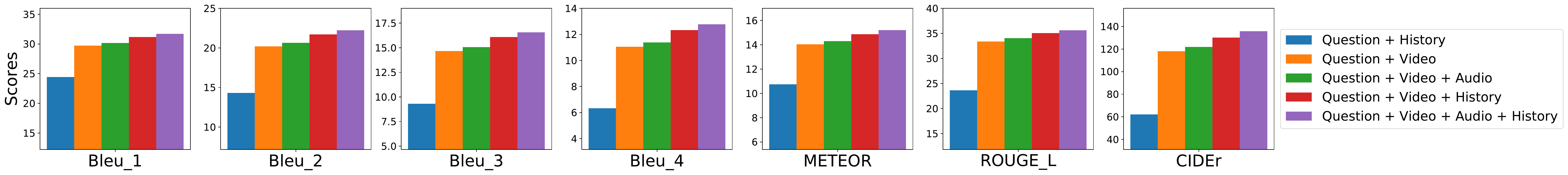}
\end{center}
\caption{Impact of different combinations of multimodal inputs on performance of \textsc{Vx2Text} for the task of audio-visual scene-aware dialog on the AVSD validation set. (Best viewed in colors.) Each modality contributes to elevate the performance, especially the video signal.}
\label{avsd_m}
\end{figure*}

\begin{figure}[!t]
\begin{center}
\includegraphics[width=0.9\linewidth]{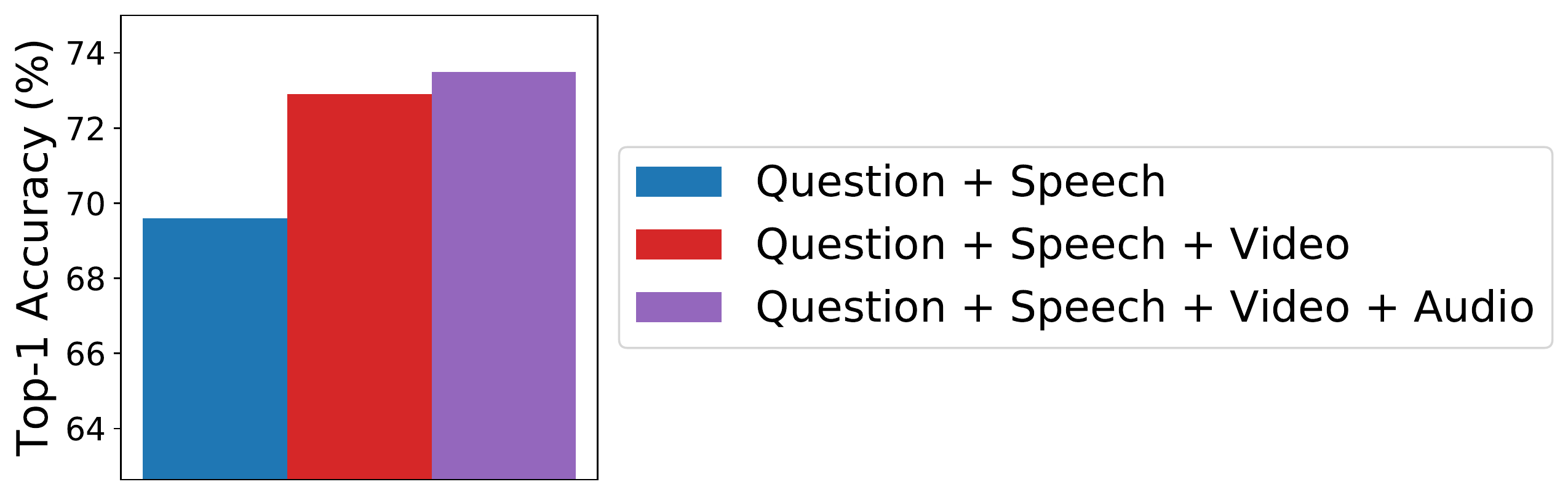}
\end{center}
\caption{Video question answering performance of \textsc{Vx2Text} on the TVQA validation set for different combinations of input modalities. (Best viewed in colors.)} 
\label{tvqa_m}
\end{figure}

We use T5-base \cite{T5} as our text transformer including the text token embedding layer, the encoder and the decoder. We use pretrained weights provided in HuggingFace~\cite{wolf2019huggingface} for initialization of the text transformer. We note that, except for these initializations, we do not use any form of pretraining and that the optimization of the model is done on each individual task using the given training set.

We use a batch size of 6 examples per GPU, and distribute the training over 32 NVidia V100 GPUs. We use Adam with a learning rate of 0.0001 to optimize our models. We train our models for 40 epochs, with the learning rate divided by 10 at the 20-th and 30-th epochs. Training on AVSD, TVQA, and TVC with our default settings takes about 12, 15, and 20 hours, respectively.

\section{Experiments}

In this section, we evaluate the effectiveness of \textsc{Vx2Text} on three distinct tasks: (1) video questions answering, (2) audio-visual scene-aware dialog, and (3) video captioning. We use three benchmark datasets: TVQA, AVSD, and TVC for these three tasks, respectively.

\subsection{Datasets and Evaluation Metrics}

\textbf{Audio-Visual Scene-Aware Dialog.} AVSD \cite{alamri2018audio} is a benchmark consisting of human dialogs describing videos in the Charades dataset \cite{Charades1}. The dialogs are in the form of 10 question-answer (QA) pairs per video. The questions are formulated by a human subject who has not observed the video. The aim of the questions is to collect as much information as possible about the content of the video. This is accomplished through a dialog with a person who has seen the video and provides detailed answers to the questions. Algorithms are evaluated on this benchmark by their ability to answer the questions in textual form. As in prior work~\cite{schwartz2019simple}, we adopt the following evaluation metrics: BLEU-\{1,2,3,4\} \cite{papineni2002bleu}, CIDEr \cite{vedantam2015cider}, METEOR \cite{banerjee2005meteor}, and ROUGE-L \cite{lin2004rouge}. We follow the common practice of training on the training split, using validation set for ablation studies, and reporting performance on the test set for comparison with the state-of-the-art. 

\textbf{Video Question Answering.} TVQA \cite{lei2018tvqa} is a dataset consisting of video clips collected from 6 TV series. Given a video clip and its corresponding speech, the goal of this task is to answer a multiple-choice question about the clip. Each video clip has 7 questions, with 5 candidate answers per question. In total, the dataset consists of 152,500 QA pairs from 21,800 clips. The speech data comes in the form of manually annotated transcripts. We use the training split to train our model and we report results on the validation set. We adopt top-1 accuracy as the standard evaluation metric.

\textbf{Video Captioning.} TVC \cite{lei2020tvr} is a recently introduced benchmark for video captioning. The TVC dataset includes the same set of videos as TVQA, but the videos are segmented into clips in a different way. We follow the protocol introduced in previous work \cite{lei2020tvr} and include the speech consisting of manual transcripts as input to our model. We adopt the following evaluation metrics: BLEU-\{1,2,3,4\} \cite{papineni2002bleu}, CIDEr\cite{vedantam2015cider}, METEOR \cite{banerjee2005meteor}, and ROUGE-L \cite{lin2004rouge}. The performance is evaluated on the validation set of the dataset.

\subsection{Assessing the Importance of Each Modality}

We begin by studying the effect of individual modalities on video-based text generation performance. We do so by training and testing our model with different combinations of inputs. Results are shown in Figure~\ref{avsd_m} for the AVSD dataset, and in Figure~\ref{tvqa_m} for the TVQA dataset. Based on these results, we observe that each modality provides a performance gain for both tasks. This is especially noticeable for the AVSD benchmark, which was specifically designed for multimodal understanding. Furthermore, note that the addition of the video modality yields a very significant gain under all metrics on AVSD compared to the version of our model relying only on textual input (question and history). This trend also holds on the TVQA dataset. Finally, we also observe that leveraging the history of previous QA pairs is highly beneficial for the performance on AVSD. This suggests that our model successfully incorporates information from previous QA pairs in the dialog. 

\begin{figure*}[t]
\begin{center}
\includegraphics[width=0.95\linewidth]{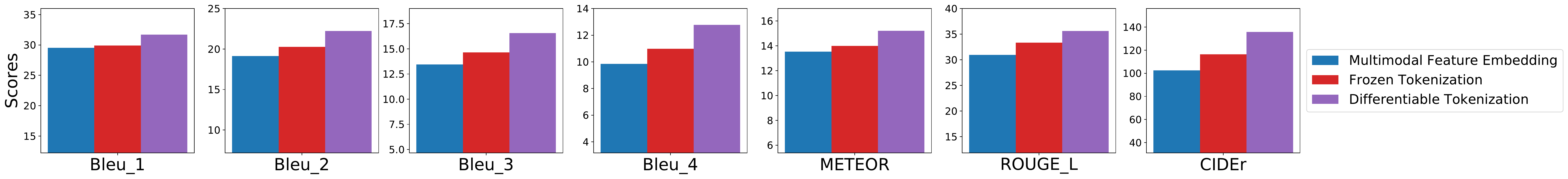}
\end{center}
\caption{Comparing performance obtained with our Differentiable Tokenization vs the baselines of Multimodal Feature Embedding or Frozen Tokenization on the AVSD validation set. (Best viewed in colors.) Differentiable Tokenization enables end-to-end training with respect to the end objective and yields the best performance.}
\label{avsd_t}
\end{figure*}

\begin{figure}[t]
\begin{center}
\includegraphics[width=0.9\linewidth]{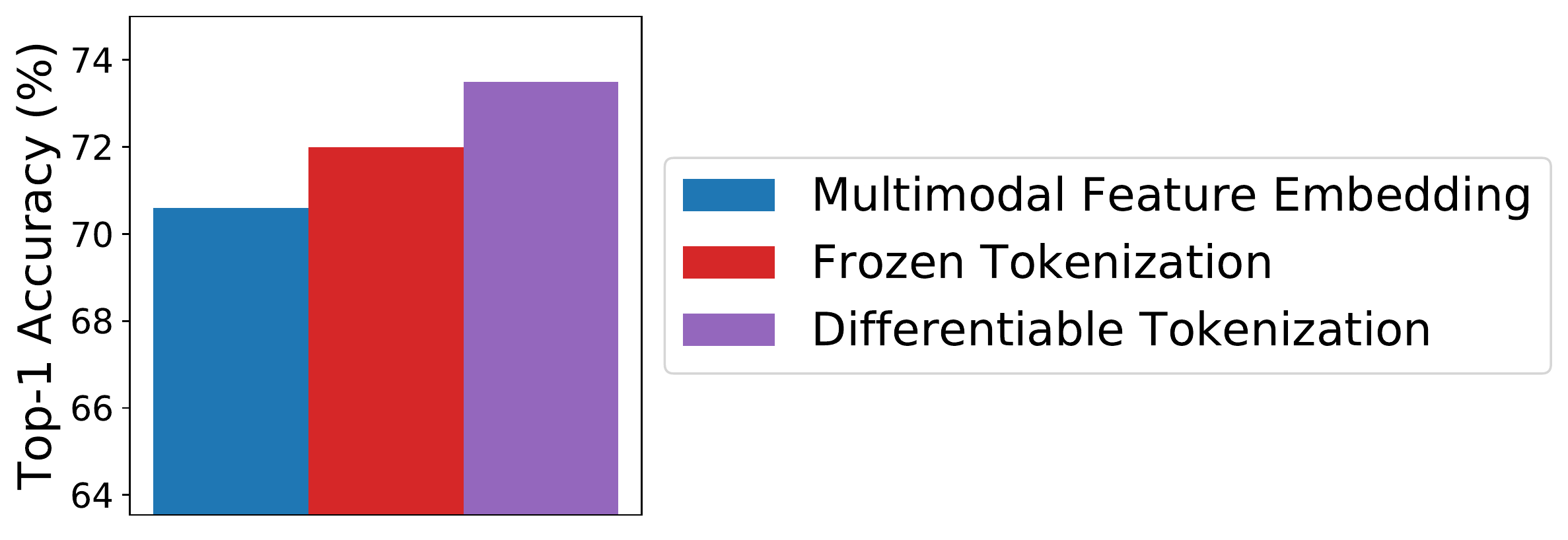}
\end{center}
\caption{Studying the effect of different modality fusion mechanisms on the QA performance of the system on the TVQA validation set. (Best viewed in colors.) Differentiable Tokenization outperforms the other schemes by large margins.}
\label{tvqa_t}
\end{figure}

\subsection{The Effect of Differentiable Tokenization}
In this section, we demonstrate the usefulness of our proposed Differentiable Tokenization scheme. 
For this purpose, we consider and test two comparative baselines. The first, named Multimodal Feature Embedding, uses a modality-specific fully-connected layer with Layer Normalization \cite{ba2016layer} to map the the continuous predictions of the audio and video classifiers into the language embedding space. This scheme is similar to the strategy implemented by the input embedding modules in HERO \cite{li2020hero} and it provides an alternative way to enable end-to-end training.



For the second baseline, we replace our Differentiable Tokenization with Frozen Tokenization, which means that only the text transformer is trained with respect to the target task, while the modality-specific networks are frozen. The results are shown in Figures~\ref{avsd_t} for AVSD and in Figure \ref{tvqa_t} for TVQA, using all available input modalities for both tasks. It can be observed that Frozen Tokenization achieves better performance than the Multimodal Feature Embedding. This by itself already provides evidence of the benefit obtained by mapping all modalities into the language space using the top predictions of the modality-specific classifiers. However, it can be noticed that Differentiable Tokenization boosts further the performance on both tasks by jointly optimizing the entire model end-to-end. 

\subsection{The Benefit of a Generative Model}
\label{sec:gen}
To show the benefits of our unified generative formulation, we present a comparison involving four models trained and evaluated on TVQA. The first model is our default \textsc{Vx2Text} model, denoted here as Generative. The second model is a discriminative version of our system obtained by removing the decoder and by attaching a classification head to the pooled embedding obtained from the encoder. This variant is trained end-to-end to predict a distribution over the five candidate answers. It is similar to the approach taken in HERO~\cite{li2020hero}, except that it uses our Differentiable Tokenization as modality fusion mechanism. As a reference, we found that our Discriminative baseline achieves performance comparable with that of HERO (without pretraining) on TVQA. 

\begin{figure}[t]
\begin{center}
\includegraphics[width=0.95\linewidth]{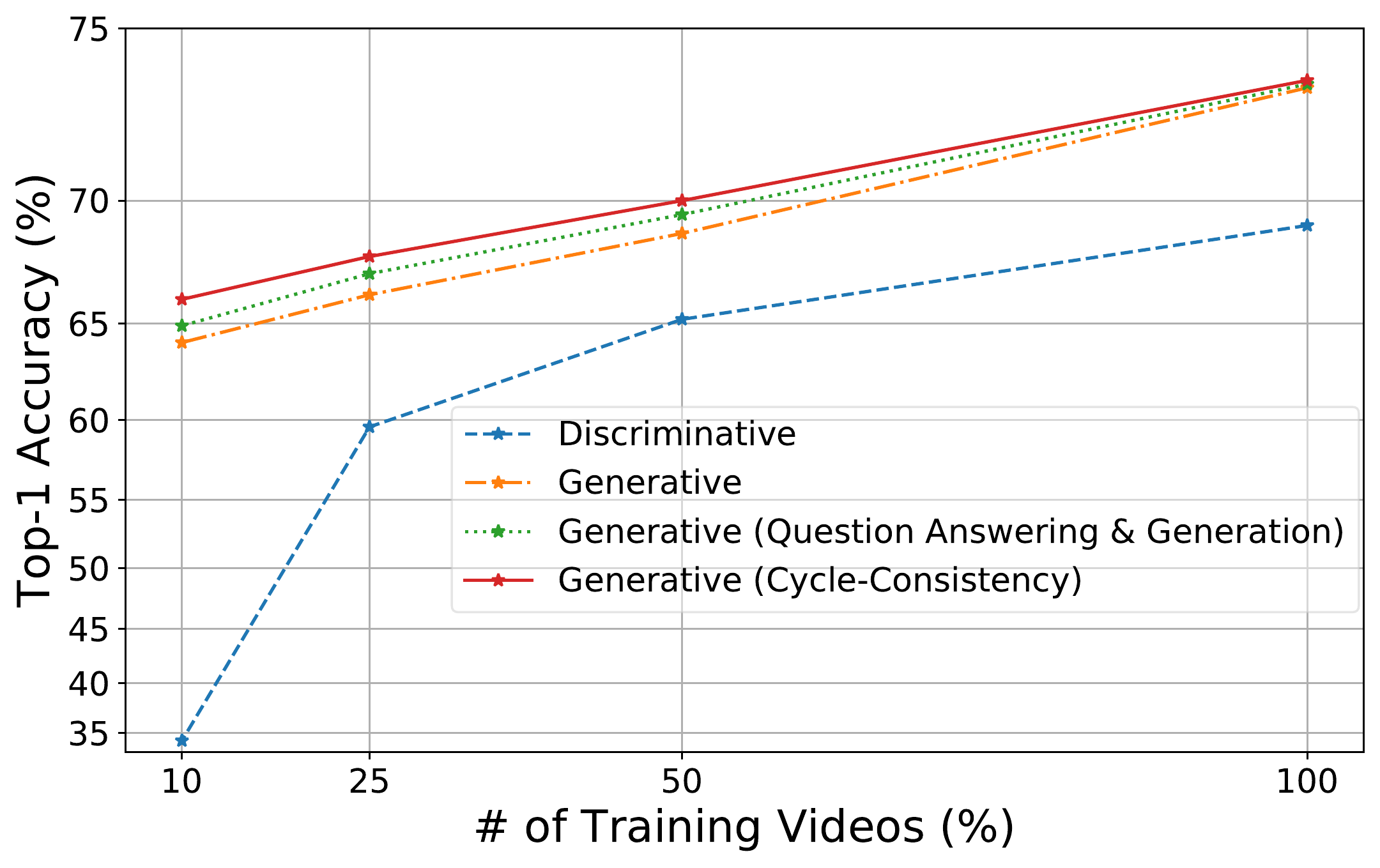}
\end{center}
\caption{Comparison between a Discriminative variant of our \textsc{Vx2Text} and the default Generative version on TVQA. The generative version achieves much higher accuracy for all training set sizes. Furthermore, the generative formulation enables multi-task learning (see ``Question Answering \& Generation'' and ``Cycle-consistency'') with the {\em same} model. This yields further improvements in accuracy, especially for small training sets.}
\label{generate}
\end{figure}

\begin{table*}[t]
  
\begin{center}
\footnotesize
\begin{tabular}{lllllllll}
\hline
Models   & Use Caption? & CIDERr  & BLEU-4 & BLEU-3 & BLEU-2 & BLEU-1 & ROUGE-L & METEOR 
\\ \hline 
MA-VDS~\cite{hori2019end}   & No      & 0.727 & 0.078 & 0.109 & 0.161 & 0.256 & 0.277 & 0.113 \\ 
Simple~\cite{schwartz2019simple}    & No      & 0.905 & 0.095 & 0.130 & 0.183 & 0.279 & 0.303 & 0.122 \\ 
\textbf{\textsc{Vx2Text} (Ours)}   & No      & \textbf{1.357} & \textbf{0.127} & \textbf{0.166} & \textbf{0.222} & \textbf{0.317} & \textbf{0.356} & \textbf{0.152} \\ 
\hline \hline
MTN~\cite{le2019multimodal}     & Yes      & 1.249& 0.128 & 0.173 & 0.241& 0.357    & 0.355 & 0.162 \\ 
MTN-TMT~\cite{li2020tmt}  & Yes      & 1.357 & 0.142 & - & -& -   & 0.371 & 0.171 \\ 
\textbf{\textsc{Vx2Text} (Ours)}    & Yes      & \textbf{1.605} & \textbf{0.154} & \textbf{0.197} & \textbf{0.260} & \textbf{0.361} & \textbf{0.393} & \textbf{0.178} \\ 
\hline
\end{tabular}
\end{center}

\caption{Comparison to the state-of-the-art on the AVSD test set with and without caption as input. Our model achieves the best results under both settings.}

\label{sota:avsd}
\end{table*}

\begin{table*}[t]
  
\begin{center}
\footnotesize
\begin{tabular}{lllll}
\hline
Models   & \# Samples for Multimodal Pretext & Val & Test
\\ \hline 
HERO~\cite{li2020hero}    & 7.6M     & 74.8 & 73.6 \\ 
\hline \hline
TVQA~\cite{lei2018tvqa}    & 0      &  67.7 & 68.5 \\ 
STAGE~\cite{lei2019tvqa+}    & 0     & 70.5 & 70.2 \\ 
HERO~\cite{li2020hero}    & 0     & 70.7 & 70.3 \\
MSAN~\cite{kim2020modality}    & 0     & 71.6 & 71.1  \\
BERT QA~\cite{yang2020bert}   & 0     & 72.4 & 72.7 \\ 
\textbf{\textsc{Vx2Text} (Ours)}    & 0      & \textbf{74.9} & \textbf{75.0}  \\ 
\hline

\end{tabular}
\end{center}

\caption{Comparison to the state-of-the-art for the task of Video Question Answering on both the validation set and the test set of TVQA. On the test set, \textsc{Vx2Text} achieves even better performance than the version of HERO that leverages 7.6M additional multimodal samples for pretraining.  Numbers represent Top-1 Accuracy (\%).}

\label{sota:tvqa}
\end{table*}

\begin{table*}[h]
  
\begin{center}
\scriptsize
\begin{tabular}{llllllllll}
\hline

Models   & \# Samples for Multimodal Pretext & \multicolumn{4}{c|}{Validation} & \multicolumn{4}{c}{Test} \\
\cline{3-10}
& &  CIDERr  & BLEU-4 &  ROUGE-L & METEOR & CIDERr  & BLEU-4 &  ROUGE-L & METEOR 
\\ \hline 

HERO~\cite{li2020hero}   & 7.6M      & 0.505 & 0.123  & 0.341 & 0.175& 0.500 & 0.124  & 0.342 & 0.176 \\ 

\hline \hline
MMT~\cite{lei2020tvr}     & 0       & 0.444 & 0.105    & 0.324 & 0.166 & 0.454 & 0.109    & 0.328 & 0.169 \\ 
HERO~\cite{li2020hero}   & 0       & 0.436 & 0.107   & 0.327 & 0.164 & 0.437 & 0.109   & 0.326 & 0.165 \\ 

\textbf{\textsc{Vx2Text} (Ours)}   & 0      & \textbf{0.482} & \textbf{0.116}  & \textbf{0.328} & \textbf{0.172} & \textbf{0.483} & \textbf{0.119}  & \textbf{0.331} & \textbf{0.174} \\ 
\hline
\end{tabular}
\end{center}

\caption{Video captioning performance of \textsc{Vx2Text} on both the validation set and the test set of TVC. Our model achieves the best performance among the methods that do not make use of additional samples for multimodal pretraining.}
\label{sota:tvc}
\end{table*}

Furthermore, to show the flexibility of our generative formulation, we include two additional variants of \textsc{Vx2Text} using multiple generative training objectives. ``Generative (Question Answering \& Generation)'' has two training objectives: one is for video question answering and the other is for video question generation. When generating questions, our model takes $Question$ as the task token $t$ and the ground-truth answer as part of the input. In such mode the system is asked to predict in a generative manner the ground-truth question from the ground-truth answer.

In ``Generative (Cycle-consistency)'', our model performs the following steps: 1) generates answer $A'$ given the ground-truth question $Q$; 2) produces question $Q''$ based on $A'$; 3) outputs answer $A''$ based on $Q''$. The final objective is a linear combination of the Question consistency $|Q''-Q|$, the Answer consistency $|A''-A|$ as well as the Question Answering and Question Generation losses. Such a multi-loss objective was originally proposed by Shah et al.~\cite{shah2019cycle} for the case of image-based QA. For details of these two baselines, please refer to our appendix.

Figure~\ref{generate} shows the performance of these four models as we vary the number of QA pairs used for training. Our \textsc{Vx2Text} model trained in a generative fashion significantly outperforms its discriminative counterpart for all training set sizes, but especially so when data is dramatically reduced. For example, the accuracy gap between Generative and Discriminative is 29.9\% (64.1\% vs 34.2\%) when using 10\% of the training data. We believe that this large performance difference comes from the beneficial commonsense knowledge stored in the text decoder. 

Moreover, our generative formulation allows \textsc{Vx2Text} to be trained with respect to multiple tasks without the need to change the architecture or add network heads. As shown in the Figure, this translates in further performance improvements. For example, with the help of Cycle-consistency, our \textsc{Vx2Text} achieves an accuracy of 66.1\% (vs the 64.1\% of Generative) when using 10\% of the data. Our \textsc{Vx2Text} trained with Cycle-consistency using only 50\% of the training samples outperforms the Discriminative model trained on the full training set (100\% of the samples).

\subsection{Comparison With the State-of-the-Art}
\label{sec:sota}
In this section we compare our single model to the state-of-the-art on the three separate benchmarks. 

\textbf{AVSD.} Our comparative results on this benchmark are shown in Table~\ref{sota:avsd}. Our \textsc{Vx2Text} significantly improves over existing methods both with and without text caption as part of the inputs. Note that the state-of-the-art MTN system~\cite{le2019multimodal} uses complex cross-modal attentional modules to fuse the information from different modalities. MTN-TMT~\cite{li2020tmt} leverages complex auxiliary losses to align the embedding spaces of MTN. However, even without text caption, which is a very strong information source, our \textsc{Vx2Text} achieves already better performance than MTN. When adding text caption to the input, the performance of our \textsc{Vx2Text} is further boosted and significantly outperforms MTN-TMT. This further demonstrates the effectiveness of our simple scheme for modality integration.

\textbf{TVQA.} Since many methods on TVQA use object/frame-level features, for a fair comparison, we include detected object categories~\cite{lei2018tvqa} as an extra modality of input for this evaluation with our \textsc{Vx2Text}. Due to the complexity of training object detectors, here we use Frozen Tokenization and leave the application of Differentiable Tokenization for future work. 

Table~\ref{sota:tvqa} shows that on TVQA our \textsc{Vx2Text} significantly outperforms all previous methods on both the validation set and the test set when training is done without additional multimodal pretext training data. On the test set, our \textsc{Vx2Text} yields an improvement of 1.4\% compared to the previous state-of-the-art, represented by the HERO system which adopts an expensive multimodal pretext training on 7.6M additional samples. As reported in~\cite{li2020hero}, this pretraining takes about 3 weeks. When both models are trained without multimodal pretext, our \textsc{Vx2Text} outperforms HERO by 4.7\%. 

\begin{figure*}[t]
\begin{center}
\includegraphics[width=0.99\linewidth]{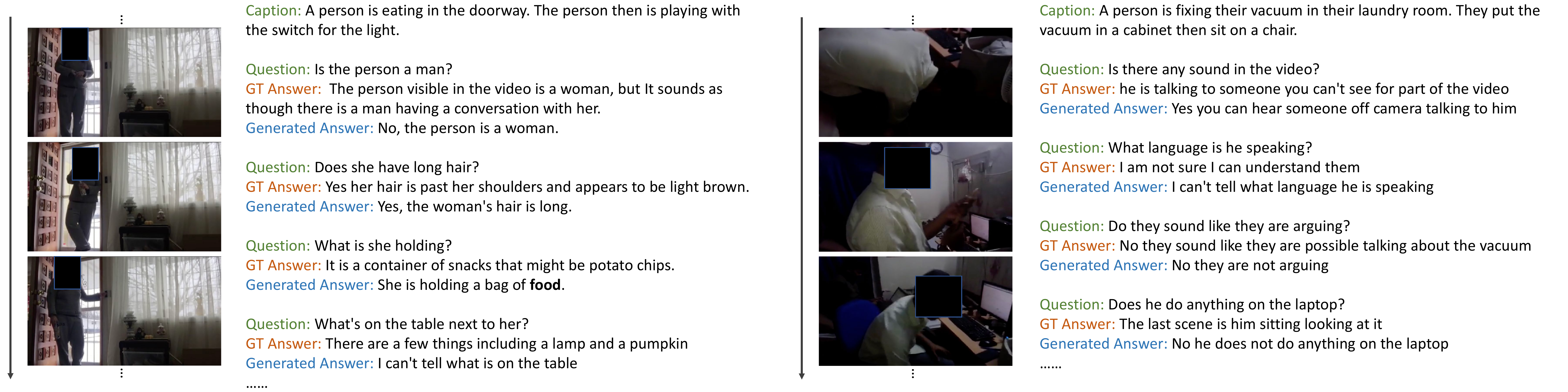}
\end{center}
\caption{Examples of generated answers for audio visual scene-aware dialog on the AVSD validation set. Our \textsc{Vx2Text} successfully responds in natural language given the multimodal inputs. Faces in the frames are artificially masked for privacy reasons.}
\label{vis:avsd}
\end{figure*}

\begin{figure*}[!t]
\begin{center}
\includegraphics[width=0.99\linewidth]{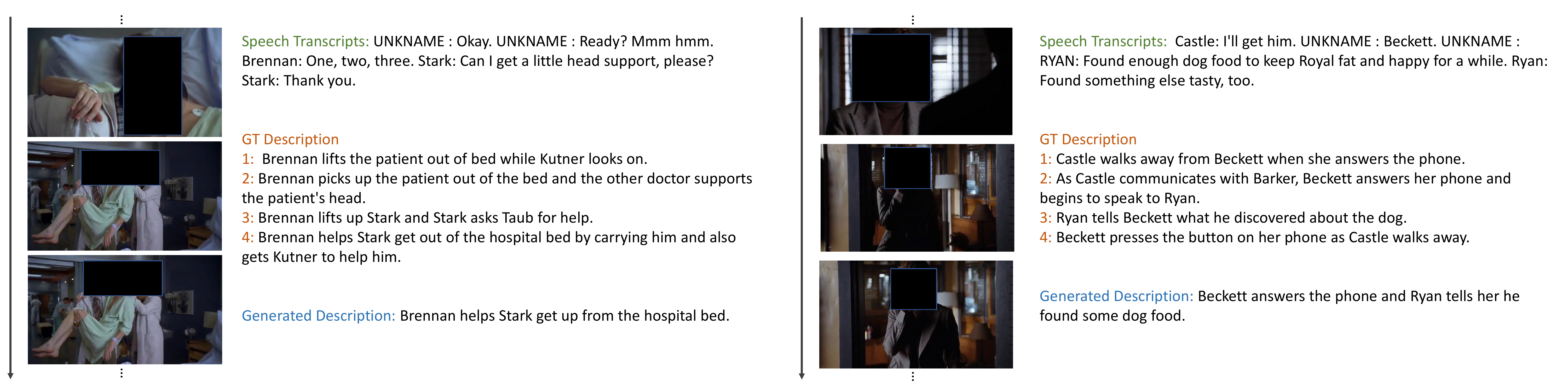}
\end{center}
\caption{Examples of textual descriptions generated by \textsc{Vx2Text} for video captioning on the TVC validation set. Our \textsc{Vx2Text} generates informative descriptions from multimodal inputs. Faces in the frames are artificially masked for privacy reasons.}
\label{vis:tvc}
\end{figure*}

\textbf{TVC.} On the captioning task of TVC our proposed \textsc{Vx2Text} significantly outperforms the state-of-the-art MMT~\cite{lei2020tvr} system. Without pretraining, HERO achieves performance comparable to that of MMT and inferior to ours. With multimodal pretraining on additional 7.6M samples (again requiring 3 weeks), HERO does only slightly better than our model. Our \textsc{Vx2Text} also shows its great generalization on the test set. Note that, as done on TVQA, even here we include object detection predictions as an input modality for our model since the methods considered in this comparison all have access to frame-level features.



\subsection{Qualitative Results}
As shown in Figures~\ref{vis:avsd} and \ref{vis:tvc}, our \textsc{Vx2Text} generates realistic natural text for both audio-visual scene-aware dialog and video captioning. It is very encouraging that although our model takes some text inputs, e.g., dialog histories or speech transcripts, the generated text does include information from other modalities. For example, as the examples in Figure \ref{vis:tvc} shows, our model successfully recognizes the actions, e.g., helping to get up or answering the phone, and even grounds the characters correctly. Please see the appendix for additional qualitative results.

\section{Conclusions}

In this work we have presented a simple unified framework to address the problem of text generation from video with additional modalities. Our approach hinges on the idea of mapping all modalities into a semantic language space in order to enable the direct application of transformer networks, which have been shown to be highly effective at modeling language problems. We have introduced a mechanism of differentiable tokenization to convert the continuous outputs of modality-specific classifiers into the language space. This renders our entire model trainable end-to-end. Our framework applied to a single architecture outperforms the state-of-the-art on three different video-based text-generation tasks.

\section*{Acknowledgments}
Thanks to Bruno Korbar, Rohit Girdhar and Fabio Petroni for their work in an early version of this project. This research was done while Xudong Lin was an intern at Facebook AI.  

{\small
\bibliographystyle{ieee_fullname}

\bibliography{egbib}
}
\clearpage
\appendix
\section{Additional Qualitative Results}
In order to understand the semantics of video and audio predictions learned through the end-to-end training process, we visualize the correlation between video/audio categories and generated words using WordCloud\footnote{https://github.com/amueller/word\_cloud}. In order to visualize the most salient words, we first remove all the stop words~\cite{nltk} from the captions generated by \textsc{Vx2Text} on the TVC test set. Then we also remove from the generated captions the words included in the corresponding input speech transcripts. By doing so, the remaining words in the generated caption have a higher chance to be derived from (or at least influenced by) the video/audio input. Finally, for each video/audio category, we consider all the (remaining) words generated by \textsc{Vx2Text} when that category is sampled and rank them according to the TF.IDF~\cite{rajaraman2011mining} score. The TF.IDF score for each word given the sampled video/audio category is regarded as the correlation between them.

\begin{figure*}[t]
\begin{center}
\includegraphics[width=0.95\linewidth]{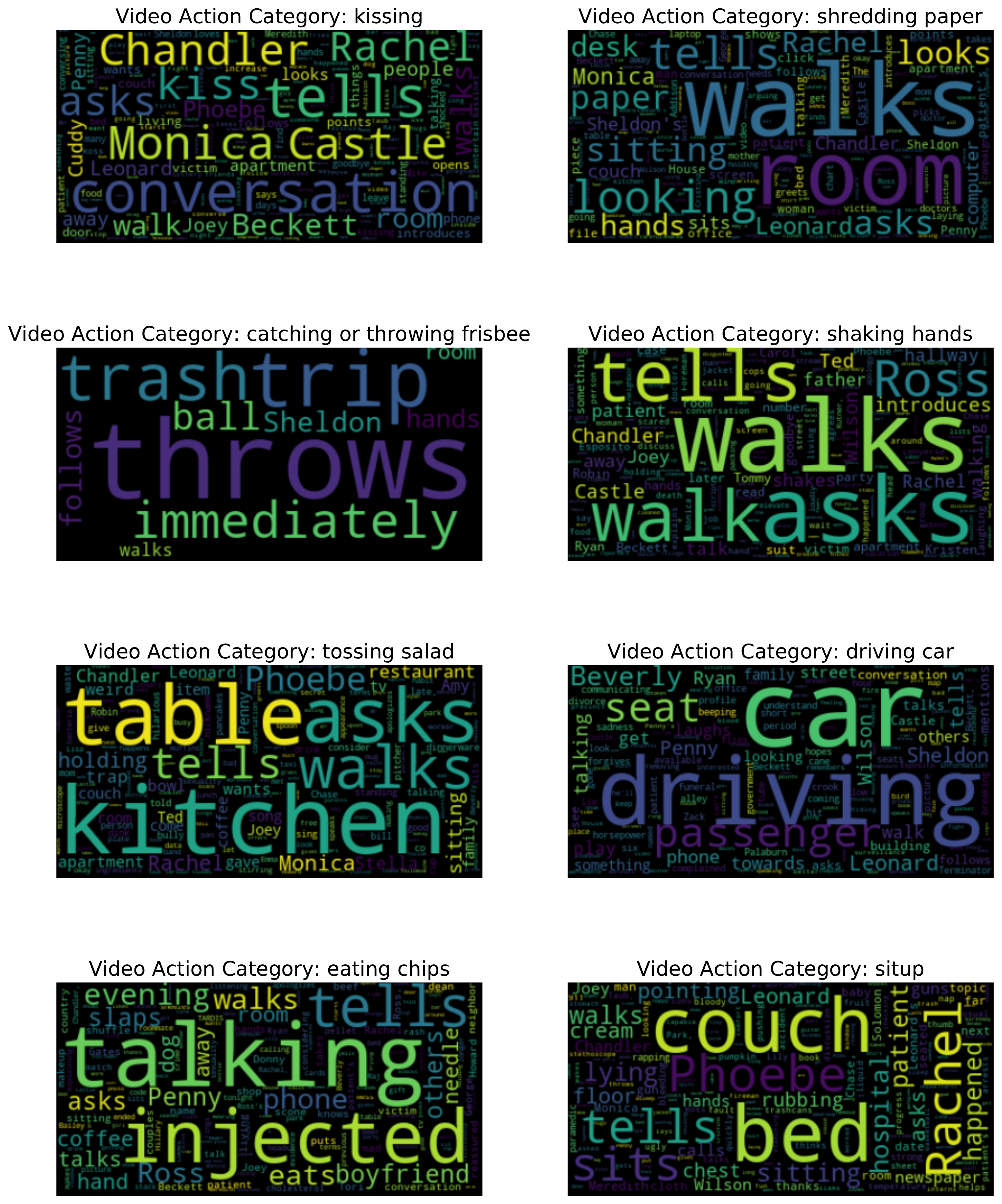}
\end{center}
\caption{Cloud visualizations of words generated by \textsc{Vx2Text} on the TVC test set conditioned on specific top action categories sampled from the video. The higher the correlation, the larger the font.}
\label{video_token}
\end{figure*}

\begin{figure*}[t]
\begin{center}
\includegraphics[width=0.95\linewidth]{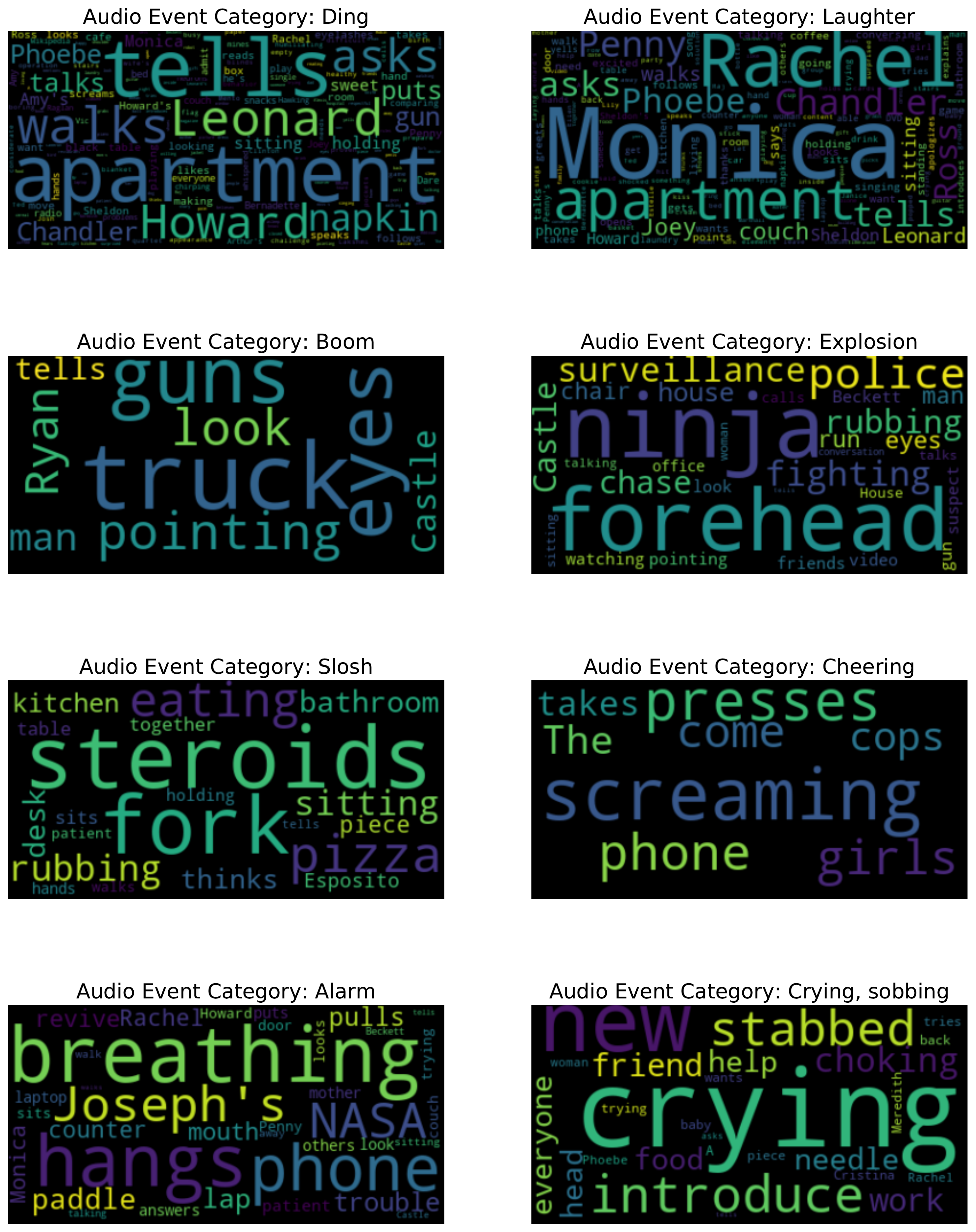}
\end{center}
\caption{Cloud visualizations of words generated by \textsc{Vx2Text} on the TVC test set conditioned on specific top sound categories sampled from the audio. The higher the correlation, the larger the font.}
\label{audio_token}
\end{figure*}

Figure~\ref{video_token} and Figure~\ref{audio_token} show the visualization of 8 categories for video actions and audio events, respectively. We observe that when the categories are general enough, their literal meaning is well maintained by the training process. For example, as shown in Figure~\ref{video_token}, kiss is one of the most correlated words to the action category ``kissing'', and driving and car are highly correlated to the action category ``driving car.'' Similarly, crying is one of the most correlated words to the audio event category ``crying, sobbing''. 

For category names that are not immediately useful for captioning, we observe interesting semantic shifts of the category names as a result of the end-to-end learning. For example, as shown in Figure~\ref{video_token}, kitchen and table are the most correlated words to the action category ``tossing salad'', while couch and bed are heavily correlated to the action category ``situp.'' Similarly, as seen in Figure~\ref{audio_token}, apartment is highly correlated to the audio event category ``ding'', while police is correlated to the audio event category ``explosion''. 

These visualizations suggest that \textsc{Vx2Text} does learn informative embeddings for video and audio categories and successfully integrates these embeddings for open-ended text generation.

\begin{figure}[!h]
\begin{center}
\includegraphics[width=0.95\linewidth]{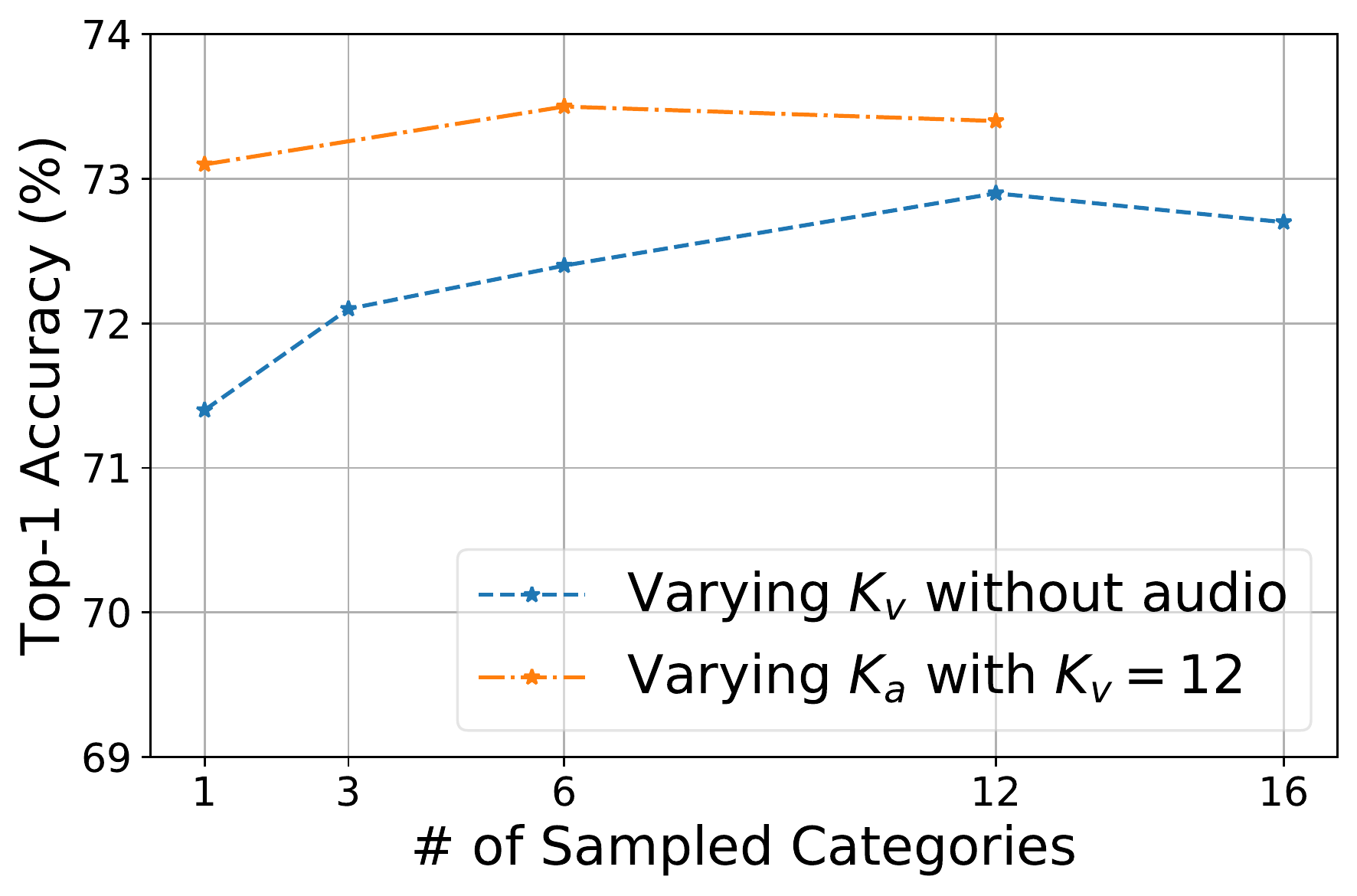}
\end{center}
\caption{Impact of different number of video categories ($K_v$) and audio categories ($K_a$) on the performance of \textsc{Vx2Text} for the task of video question answering using the validation set of TVQA.}
\label{sens}
\end{figure}

\section{Hyper-Parameter Tuning: $K_v$ and $K_a$}
In Figure~\ref{sens} we report TVQA validation results obtained by varying the number of video categories ($K_v$) and the number of audio categories ($K_a$) provided as input to the encoder. In order to reduce the computational cost of this experiment, we adopt a two-stage tuning strategy. We first exclude audio from the input to \textsc{Vx2Text} and only explore the best values for $K_v$. Based on these results, we fixed $K_v=12$ and subsequently searched for the optimal $K_a$ by adding audio to the input. It can be seen that $K_a=6$ produces the best performance. We leave the joint optimization of both hyper-parameters for future work. We also leave the hyper-parameter tuning on the embedding dimensions of video/audio categories for future work. We simply use $13\times 768$ and $15\times 768$ for video and audio categories, respectively, since the category names contain a maximum number of text tokens of 13 and 15, respectively. We pad short category names with text padding tokens.

\section{Details about Generative Training Variants}
For ``Generative (Question Answering \& Generation)'' and Cycle-consistency, we simply average (with equal weights) all the losses in each formulation as the final training objective. The losses for Question Generation, Answer-Consistency, and Question Consistency are all as described in Equation 6. For Cycle-consistency, we use greedy search to decode question or answer sentences during training. Note that for simplicity, we do not apply the Gating Mechanism and Late Activation as described in \cite{shah2019cycle} but use all the generated questions from the first epoch for Cycle-consistency training.

\end{document}